\documentclass{article}
\usepackage{graphicx}

\usepackage[T1]{fontenc}
\usepackage{lmodern} 
\usepackage{wrapfig}

\usepackage{xcolor}
\usepackage{enumitem}
\usepackage{hyperref}

\hypersetup{
  colorlinks=true,
  linkcolor=black,
  citecolor=black,
  urlcolor=blue!60!black
}

 \usepackage{neurips_2025}
\usepackage{graphicx}
\usepackage{float}
\usepackage[utf8]{inputenc} 
\usepackage[T1]{fontenc}    
\usepackage{hyperref}       
\usepackage{url}            
\usepackage{booktabs}       
\usepackage{amsfonts}       
\usepackage{nicefrac}       
\usepackage{microtype}      
\usepackage{xcolor}         

\title{Video Models Start to Solve Chess, Maze, Sudoku, Mental Rotation, and Raven' Matrices}

\author{%
  Hokin Deng \\
  Carnegie Mellon University\\
  \texttt{hokind@andrew.cmu.edu} \\
    \AND Source Code: \href{https://github.com/hokindeng/VMEvalKit}{\texttt{github.com/hokindeng/VMEvalKit}\quad \href{https://grow-ai-like-a-child.com/video-reason/}{Results Page}}
}

\begin{document}

\maketitle

\begin{abstract}
We show that video generation models could reason now. Testing on tasks such as chess, maze, Sudoku, mental rotation, and Raven’s Matrices, leading models such as Sora-2 achieve >60\% success rates. We establish a robust experimental paradigm centered on the "Task Pair" design. We build a code framework, with 39 models available already, that supports this paradigm and allows for easy scaling—users can add models and tasks efficiently. We show our automated evaluation strongly correlates with human judgment, and therefore this paradigm is highly scalable. We see an opportunity, given the availability of our paradigm, to do reinforcement learning for improving reasoning in video models.
\end{abstract}

\begin{figure}[H]
  \centering
  \begin{minipage}{\linewidth}
\includegraphics[width=\linewidth]{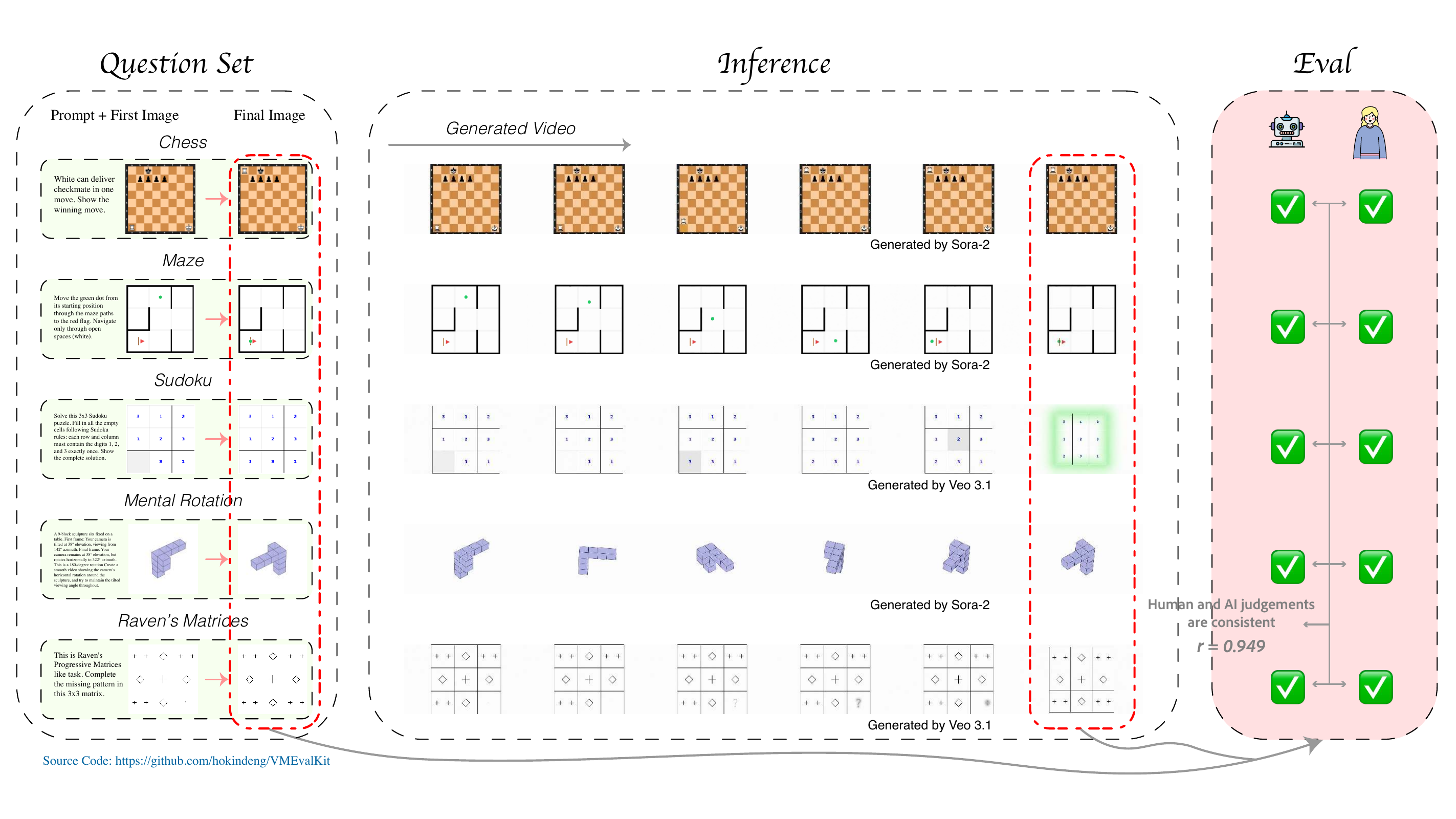}
  \caption{Representative examples of video reasoning across diverse cognitive tasks. We show 6-frame temporal sequences from videos. The robust correlation (r = 0.949, p < 0.001) between human and automated GPT-4o evaluations demonstrates the reliability of our evaluation framework for assessing reasoning in video generation models. }
  \label{fig:teaser}
  \end{minipage}
\end{figure}

\section{Introduction}

The rapid evolution of video generation models has transformed our ability to synthesize realistic, high-fidelity videos from text descriptions. Systems like Sora-2 \citep{openai_sora2_2025}, Veo-3 \citep{deepmind_veo3_2025}, Runway Gen-3 \citep{runway_gen3_2024}, and Luma Dream Machine \citep{luma_dream_machine_2024} can now generate videos that are visually indistinguishable from human-created content, depicting complex scenes with coherent motion, consistent objects, and plausible physics. These advances have been driven by scaling up training data, model parameters, and computational resources, coupled with architectural innovations. However, a fundamental question remains largely unexplored: \textbf{Can these models reason?} 

Visual reasoning—the capacity to understand, manipulate, and solve problems through visual representations—represents a qualitatively different challenge from photorealistic synthesis. While generating a realistic video of a chess game or a person solving a Sudoku puzzle requires learning statistical patterns of motion and appearance, \textbf{actually solving} these problems requires understanding the underlying rules, constraints, and logical relationships that govern valid solutions \citep{wiedemer2025video_zero_shot}. The distinction is critical: a model that has learned to mimic the visual appearance of problem-solving without understanding the problem itself cannot reliably generate correct solutions, nor can it adapt to novel problem instances requiring genuine reasoning \citep{deng2025knowing}.

\begin{figure*}[h]
\centering
\includegraphics[width=0.8\textwidth]{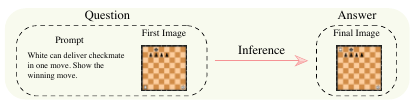}
\caption{Each of our question unit includes three components: (1) an initial image showing the unsolved problem, (2) a text instruction describing the task, and (3) a final image showing the correct solution. During inference, models are only given the initial image and the instruction prompt. The final image is withheld and used solely for evaluation.}
\label{fig:questionset}
\end{figure*}

Evaluating reasoning in video models isn’t straightforward. Our paradigm provides (1) an initial image showing the unsolved problem, (2) a text instruction prompt explaining what to do, (3) a final image showing the correct solution (See Figure \ref{fig:teaser} and \ref{fig:questionset}). In inference, we only show (1) and (2) to the video models and (3) is reserved for evaluation. We build \href{https://github.com/hokindeng/VMEvalKit} {VMEvalKit}, a flexible open-source code framework for our experimental paradigm. Our codebase could (i.) create tasks for testing reasoning at scale, (ii.) run inferences of video models on the tasks at scale, and (iii.) do AI-powered evaluation at scale. It's also modularized which the users could add more tasks and models on their own easily. Now our tasks include chess, maze, Sudoku, mental rotation, and Raven's Matrices. The code framework now provides inferences for 39 video models spanning 11 model families. 

We ran the \href{https://github.com/hokindeng/VMEvalKit} {VMEvalKit} evaluation on 450 tasks across six cutting-edge video generation models: OpenAI Sora 2, Google Veo 3.0 and 3.1, Runway Gen4 Turbo, WaveSpeed WAN 2.2, and Luma Ray 2. Each model was tested on 75 tasks covering five reasoning tasks—chess, mazes, Raven’s matrices, mental rotation, and Sudoku. Performance was measured using both human ratings and GPT-4o-based automated scoring, with strong agreement between the two (Pearson r = 0.949, Cohen’s K = 0.867), confirming the reliability of automated evaluation. Sora 2 stood out with a 68\% overall success rate, particularly excelling in maze navigation (87\%), Raven’s matrices (73\%), and chess (73\%). Veo 3.0 followed at 47\%, achieving a perfect score on Sudoku tasks, while Veo 3.1 landed at 35\%. The remaining models struggled: Gen4 Turbo reached 24\%, WAN 2.2 hit 11\%, and Luma Ray 2 scored just 1\%. Among task types, Sudoku proved the easiest (57\% average success), while mental rotation and chess were the most challenging (around 11\%). We see all of our inference results from the video models, together with the prompt and question images, here at the \href{https://grow-ai-like-a-child.com/video-reason/}{\textbf{Results Page}}

We argue that we have observed strong evidences that the video models have emerged reasoning abilities. There are important research opportunities on the horizon. First, it enables reinforcement learning: clear success signals (e.g., solved maze or legal Sudoku) make it possible to fine-tune models toward more consistent reasoning. Second, we need mechanistic interpretability to uncover how models represent rules, track state, and reason across steps. Third, functional decomposition—breaking tasks into sub-reasoning faculties—can help isolate where reasoning fails and how to improve it. Together, these works shall move us to a new paradigm-shift in AIs.

\section{Related Works}

\subsection{Reasoning in Language Models}

The emergence of reasoning capabilities in large language models has been extensively studied across multiple domains. Chain-of-thought prompting demonstrated that language models can solve complex reasoning tasks by generating intermediate reasoning steps, with performance scaling dramatically with model size \citep{wei2022cot}. This has been extended to multi-step reasoning in mathematics, logical inference, and strategic planning \citep{cobbe2021gsm8k, creswell2022selectioninference, yao2023tot}.

Recent work has focused on evaluating and improving reasoning through specialized benchmarks. GSM8K evaluates mathematical problem-solving, while BIG-Bench assesses diverse reasoning capabilities including logical deduction, causal reasoning, and analogical thinking \citep{cobbe2021gsm8k, srivastava2023bigbench}. MMLU provides comprehensive evaluation across 57 subjects requiring factual knowledge and reasoning. Process reward models have shown that rewarding correct reasoning steps rather than just final answers substantially improves reliability \citep{hendrycks2021mmlu, lightman2023prm}. However, these advances remain primarily text-based, with limited exploration of reasoning in visual and video domains.

\subsection{Video Generation Evaluation}

\begin{figure*}[t]
  \centering
  \includegraphics[width=0.8\textwidth]{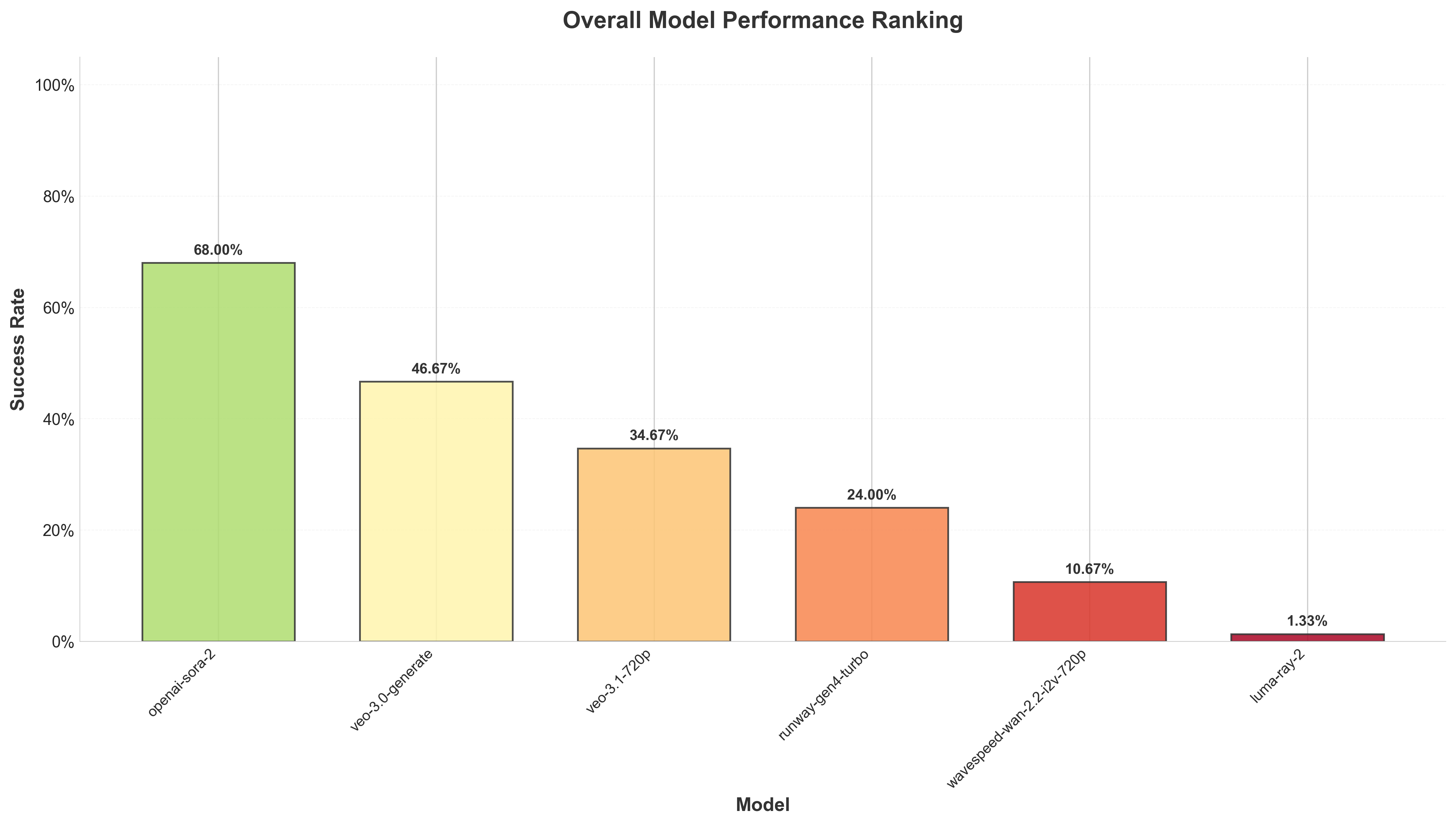}
  \caption{Success rates on all 5 tasks showing clear performance hierarchy across models.}
  \label{fig:overall}
\end{figure*}

Community “arena” efforts aggregate large-scale pairwise preferences and formalize open evaluation with Elo-style ranking \citep{artificialanalysis_vga_2025,arena_open_eval_2025}. Sometime industries also published reflection memos \citep{google_gemini_share_2025}. Dimensioned suites such as \emph{VBench} decompose quality into text relevance, temporal coherence, style, and related facets \citep{vbench_cvpr2024}, while \emph{Video-Bench} pushes closer agreement with human raters \citep{video_bench_human_aligned}. Other measures include learning-based metrics, Fréchet Video distance, and Kernel Video distance. \citep{mantisscore, fvmd1_blog}. Benchmarks also focus on specific faculties such as temporal and metamorphic robustness \citep{chronomagic_bench}, physics/world-model fidelity \citep{worldmodelbench, physics_iq_benchmark}, and so on \citep{survey_2024_video_eval}. One recent paper argues that Veo 3 exhibits broad zero-shot perception and reasoning—spanning segmentation, edge detection, editing, physical/affordance understanding, tool-use simulation, and simple maze/symmetry reasoning—suggesting video models are trending toward unified, generalist vision foundation models \citep{wiedemer2025video_zero_shot}. 

\subsection{Visual Cognition Evaluation}

\begin{figure*}[t]
  \centering
  \includegraphics[width=0.7\textwidth]{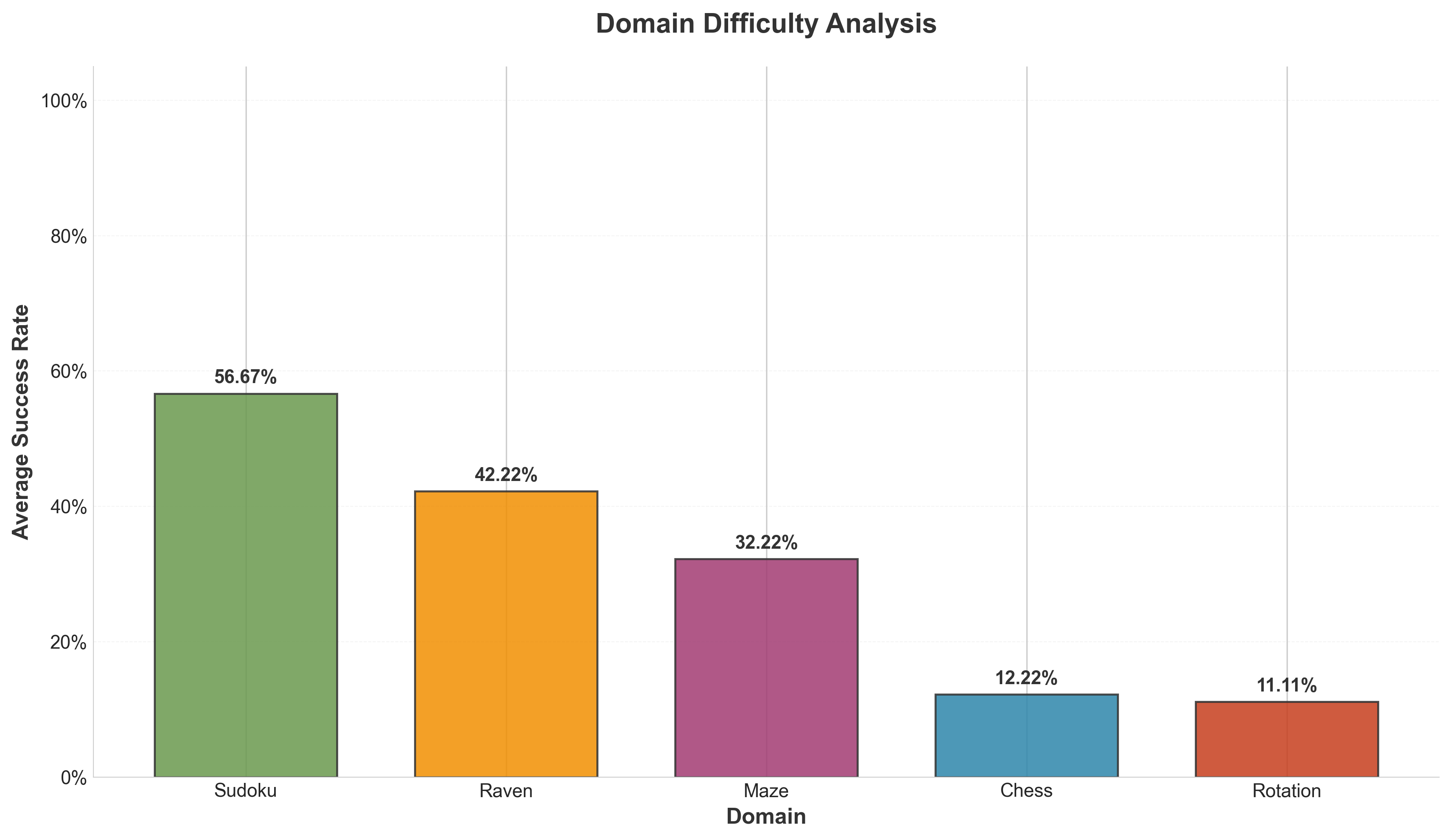}
  \caption{Average success rates by reasoning domain reveal distinct level of challenges.}
  \label{fig:difficulty}
\end{figure*}

Most of the visual cognition evaluation for AI systems are done in MLLMs. One recent project used human developmental trajectory to assess visual cognition in MLLMs \citep{li2024core, luo2025philosophical}. Other works focus on particular visual cognitive faculties such perspective taking and theory of mind \citep{gao2024vision, wang2025vision, abrini2025proceedings}, mechanical reasoning \citep{sun2024probing}, perceptual constancy \citep{sun2025probing}, gaze understanding \citep{zhang2025can}, physical conservation \citep{luo2024vision}, cognitive control \citep{luo2025machine}, and so on \citep{lievaluating, luo2025rethinking, deng2025reinforcement}. 

\section{Methods}

\subsection{Video Inference}
Our \href{https://github.com/hokindeng/VMEvalKit} {VMEvalKit} provides to evaluate 39 text-to-video models through a unified inference framework supporting both commercial APIs and open-source implementations. Our system dynamically loads model wrappers from a centralized catalog containing detailed configurations for 9 model families:
\begin{itemize}[leftmargin=*]
\item Commercial API Models: OpenAI Sora (sora-2, sora-2-pro), Google Veo (3.0-generate, 3.1 via WaveSpeed proxy), Runway ML (gen4-turbo, gen4-aleph, gen3a-turbo), Luma Dream Machine (ray-2, ray-flash-2), WaveSpeed WAN (2.1/2.2 variants with 480p/720p/5B configurations).

\item Open-Source Models: LTX-Video (13B-distilled, 13B-dev, 2B-distilled), HunyuanVideo-I2V, VideoCrafter2-512, DynamiCrafter (256p/512p/1024p variants).
\end{itemize}

Each model implements sophisticated image preprocessing to ensure integrity and homogeneity: 
\begin{itemize}[leftmargin=*]
  \item Resolution Standardization: Input images undergo automatic padding/resizing to model-specific requirements (Sora: 1280×720/720×1280, VEO: 16:9/9:16 with padding, Runway: 1280×768/768×1280 letter boxing)
  \item Color Space Conversion: Automatic RGB conversion with neutral gray padding (128,128,128) to prevent harsh borders
  \item Aspect Ratio Management: Scale-preserving letter boxing where images fit within target dimensions maintaining aspect ratio, then center-padded to exact specifications
  \item Format Standardization: Base64 encoding for API models, tensor conversion for local models, MIME type validation (image/jpeg, image/png, image/webp)
  \item Inference Parameters: During inference, all models are run using standardized settings: each video is generated with an 8-second duration, a temperature of 0.7 is applied where the model allows for sampling control, and a fixed random seed of -1 is used to promote variability across generations. Resolution handling is streamlined through automatic padding to meet each model’s input requirements.
\end{itemize}

\subsection{Task Generation}

\begin{figure*}[t]
  \centering
  \includegraphics[width=0.9\textwidth]{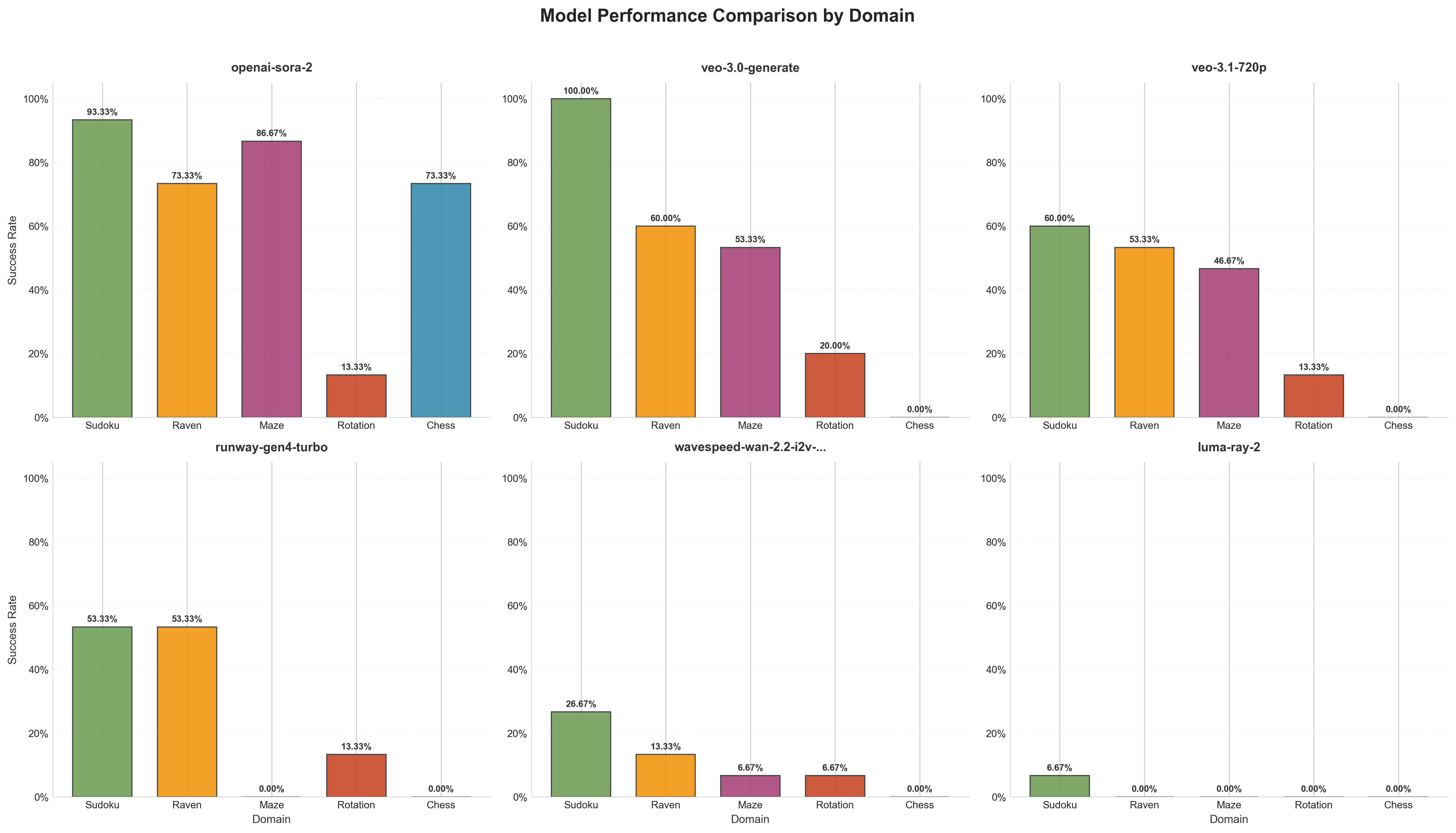}
  \caption{Individual model performance across all reasoning domains.}
  \label{fig:model_comparison}
\end{figure*}

The \href{https://github.com/hokindeng/VMEvalKit}{VMEvalKit} now provides parameterized ways to generating the following 5 tasks. For our experiments, we first have used \href{https://github.com/hokindeng/VMEvalKit}{VMEvalKit} task sets to generate several batches of tasks. Then, we use human labor to manually go through each task to ensure the task correctness and integrity. Upon our inspection, all the tasks generated by the \href{https://github.com/hokindeng/VMEvalKit}{VMEvalKit} parameterized methods are correct and good-to-go for testing the models.

\textbf{Chess Reasoning}: Mate-in-1 puzzles using a self-contained generator producing 100+ verified positions through back-rank mates, queen corner patterns, and tactical combinations. Each task shows an initial board position and requires demonstrating the winning move sequence.

\textbf{Maze Navigation}: 3×3 grid mazes generated via Kruskal's algorithm with professional rendering. Tasks display a start position (green dot) and goal (red flag), requiring path-finding demonstrations from start to finish.

\textbf{Raven Progressive Matrices}: 3×3 pattern completion puzzles testing abstract visual reasoning. Our RPM generator creates rule-based patterns (shape progression, rotation, color sequences) with systematic difficulty control.

\textbf{3D Mental Rotation}: Voxel-based structures (8–9 cubes) rendered from different camera viewpoints with 180° horizontal rotations and 20–40° elevation angles. Tasks require demonstrating spatial understanding through smooth camera transitions.

\textbf{Sudoku Solving}: Simplified 3×3 grids with exactly one missing number, testing logical deduction capabilities. Each puzzle has a unique solution following standard Sudoku constraints.

All tasks follow a consistent (first\_frame, final\_frame, prompt) format enabling standardized evaluation across domains. See Appendix~\ref{sec:detailed_task} for details. 

\section{Evaluation}

Our \href{https://github.com/hokindeng/VMEvalKit}{VMEvalKit} provides infrastructure for both human and AI automated evaluation of the results. 
\begin{itemize}[leftmargin=*]
\item Automated Evaluation: Vision-language model (GPT-4o) assessment comparing generated video final frames against ground truth solutions. The evaluator receives first frames, expected final frames, and actual video outputs, providing 1--5 correctness scores with explanations. Evaluation prompts are task-specific with detailed scoring criteria.
\item Human Evaluation: Expert annotators using a Gradio interface assess video generation quality and reasoning correctness. The system supports session resumption and consistent scoring across multiple evaluators.
\end{itemize}
Both methods use identical 5-point scales where scores 4--5 indicate successful reasoning demonstration, enabling statistical comparison of automated vs. human assessment reliability. This has been stated clearly in the prompts for both humans and AIs.

\section{Experiments}

\subsection{Overall Results}

We conduct comprehensive evaluation on 6 representative models: OpenAI Sora-2, Google Veo 3.0/3.1, Runway Gen4-Turbo, WaveSpeed WAN 2.2, and Luma Ray-2. Each model generates videos for 75 reasoning tasks (15 per domain) totaling 450 evaluations. All models use 8-second generation duration with task-appropriate prompts. OpenAI Sora-2 achieves the highest success rate at 68.0\% (3.853/5 average score), followed by Google Veo 3.0 (46.7\%), Veo 3.1 (34.7\%), Runway Gen4 (24.0\%), WaveSpeed (10.7\%), and Luma (1.3\%). This establishes a clear performance hierarchy across different model architectures and training approaches (see Figure~\ref{fig:overall}). 

Reasoning domains exhibit distinct difficulty profiles. Sudoku emerges as most tractable (56.7\% average success), followed by Raven matrices (42.2\%), Maze navigation (32.2\%), Chess solving (12.2\%), and 3D Mental Rotation (11.1\%). This ranking reflects the varying cognitive reasoning demands of each reasoning type for video models to solve (see Figure~\ref{fig:difficulty}). Superior models show consistent performance across domains, while weaker models exhibit domain-specific failures. This  reveals both general reasoning capabilities and domain-specific strengths and weaknesses. Notably, Veo 3.0 achieves very high Sudoku performance but fails Chess reasoning, indicating specialized rather than general reasoning capabilities (see Figure~\ref{fig:model_comparison}). 

The failure cases exhibit clear and often idiosyncratic patterns, but we currently lack an idea of principled framework to systematically analyze or interpret them. For now, we present some examples in the Appendix~\ref{sec:failure} and all examples in \href{https://grow-ai-like-a-child.com/video-reason/}{Results Page}. We strongly encourage readers to explore them, as they may reveal valuable insights for future investigation.

\subsection{Solving Chess}

\begin{figure*}[h]
  \centering
  \includegraphics[width=1\textwidth]{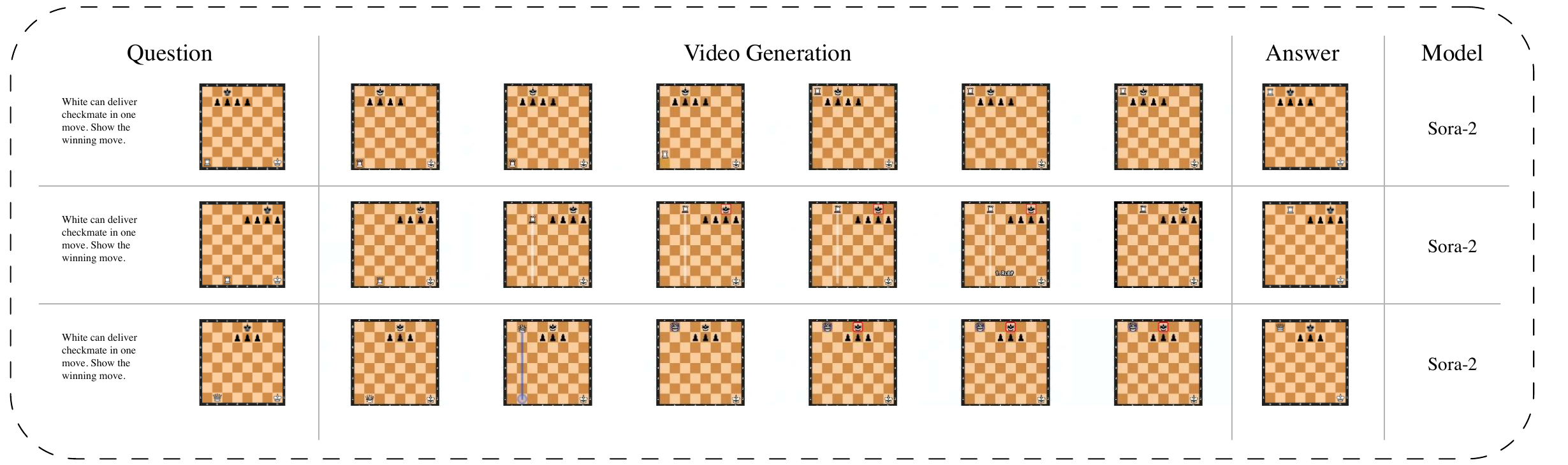}
  \caption{Example runs from video models trying to solve our chess problems. \href{https://grow-ai-like-a-child.com/video-reason/}{Results Page}}
  \label{fig:chess} 
\end{figure*}

Only OpenAI Sora-2 showed strong chess reasoning (see Figure~\ref{fig:chess}), succeeding on 73\% of tasks (11/15), especially with back-rank mates (100\%) and queen-corner mates (85\%). It failed on all knight-fork problems and multi-piece tactics, revealing limits in complex patterns. All other models—Veo 3.0/3.1, Runway Gen4-Turbo, WaveSpeed WAN 2.2, and Luma Ray-2—scored 0\%, either making legal but tactically useless moves or showing basic rule confusion. Overall, chess remain extremely challenging, with a total success rate of just 12.22\% across 90 evaluations (see Figure~\ref{fig:model_comparison}).

\subsection{Solving Maze}

\begin{figure*}[h]
  \centering
  \includegraphics[width=1\textwidth]{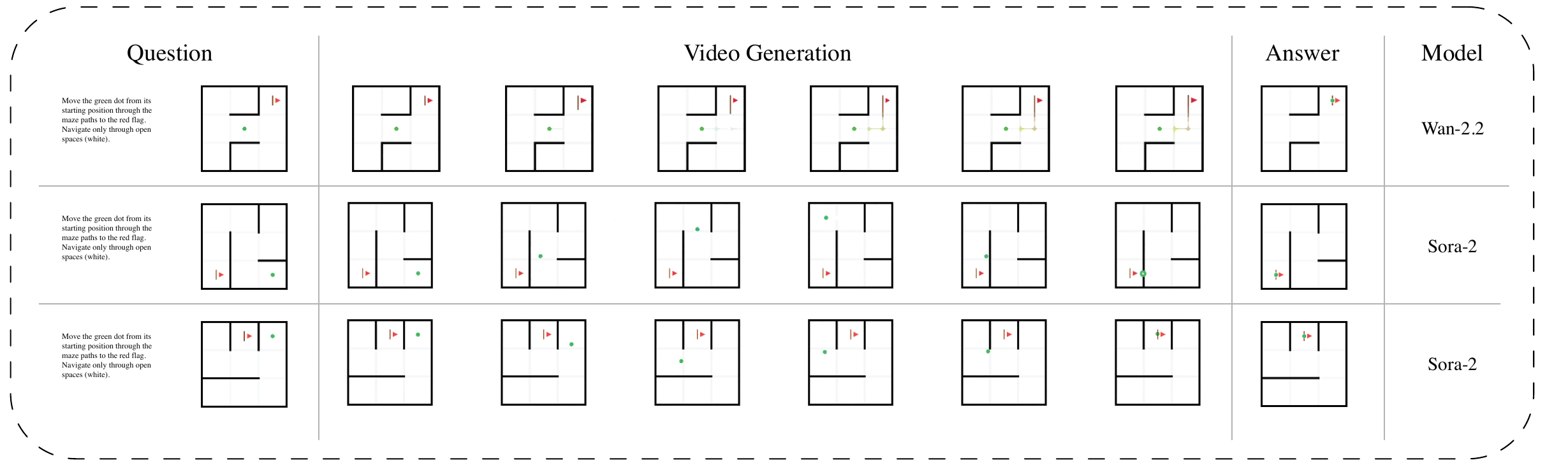}
  \caption{Example runs from video models trying to solve our maze problems. \href{https://grow-ai-like-a-child.com/video-reason/}{Results Page}}
  \label{fig:maze} 
\end{figure*}

Overall success was low, while video models do solve mazes (see Figure~\ref{fig:model_comparison} and Figure~\ref{fig:maze}). Sora-2 was the only consistently strong model, showing reliable backtracking and stable green-dot tracking. Veo 3.0/3.1 were middling (~half of tasks), better on simple, linear layouts and weaker on branching or backtracking. WaveSpeed WAN 2.2 managed a rare win on the easiest maze, while Runway Gen4-Turbo and Luma Ray-2 failed across the board. 

\subsection{Solving Sudoku}

\begin{figure*}[h]
  \centering
  \includegraphics[width=1\textwidth]{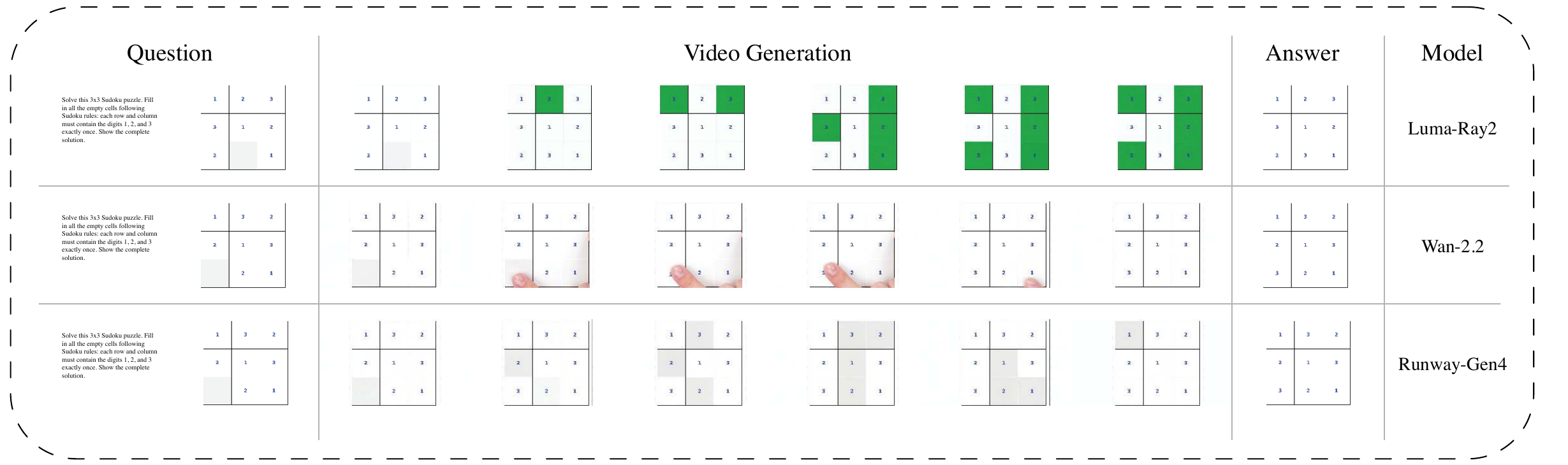}
  \caption{Example runs from video models trying to solve our Sudoku problems. \href{https://grow-ai-like-a-child.com/video-reason/}{Results Page}}
  \label{fig:sudoku} 
\end{figure*}

It turns out that Sudoku might be the most tractable domain overall, as most of the models are able to solve some (see Figure~\ref{fig:sudoku}). Veo 3.0 was almost flawless, and the same for Sora-2, with one isolated miss due to a final-frame inconsistency. Veo 3.1 and Runway Gen4-Turbo were middling—good on simpler fills, weaker when constraints interacted. WaveSpeed WAN 2.2 was weak, and Luma Ray-2 just solved one (see Figure~\ref{fig:model_comparison}).

\subsection{Solving Mental Rotation}

\begin{figure*}[h]
  \centering
  \includegraphics[width=1\textwidth]{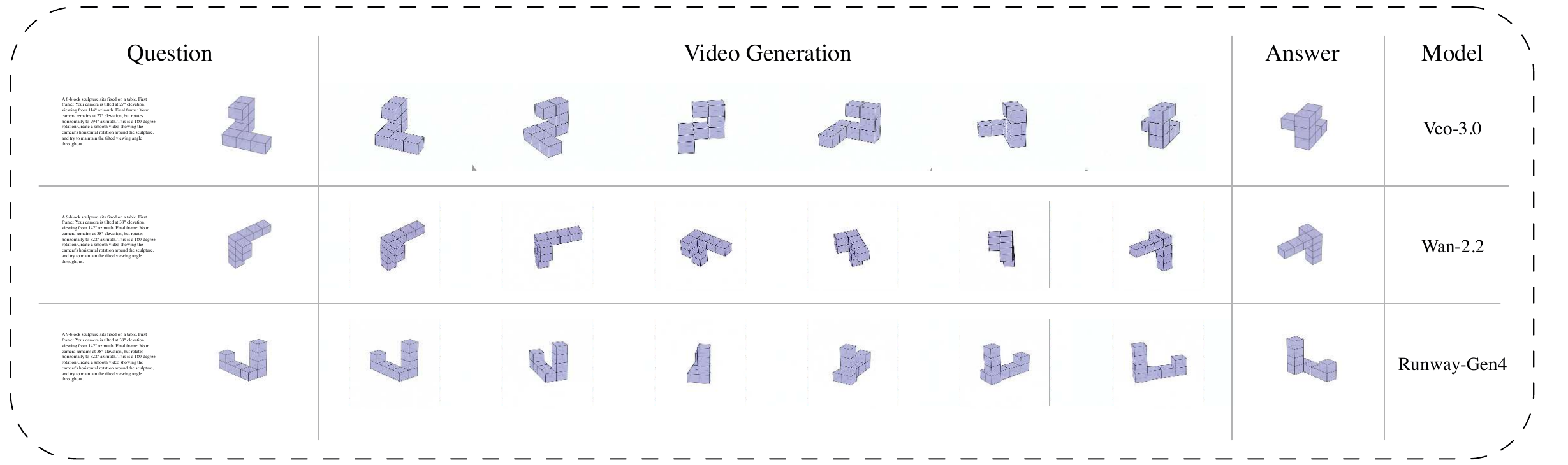}
  \caption{Example runs from video models trying to solve our Sudoku problems. \href{https://grow-ai-like-a-child.com/video-reason/}{Results Page}}
  \label{fig:rotation} 
\end{figure*}

Mental rotation seems to be a very hard problem, but weirdly video models sometimes do maintain consistency and are able to solve the rotation problems. Some failures are due to de-generation of the transformation, while other failures include not rotating to the right angle that's required in the prompt (see Figure~\ref{fig:rotation}). Veo 3.0 led but only on the simplest shapes and basic 180° turns; Sora-2, Veo 3.1, and Runway Gen4-Turbo managed occasional wins on easy layouts; WAN 2.2 rarely succeeded, and Luma Ray-2 failed entirely (see Figure~\ref{fig:model_comparison}).

\subsection{Solving Raven's Matrices}

\begin{figure*}[h]
  \centering
  \includegraphics[width=1\textwidth]{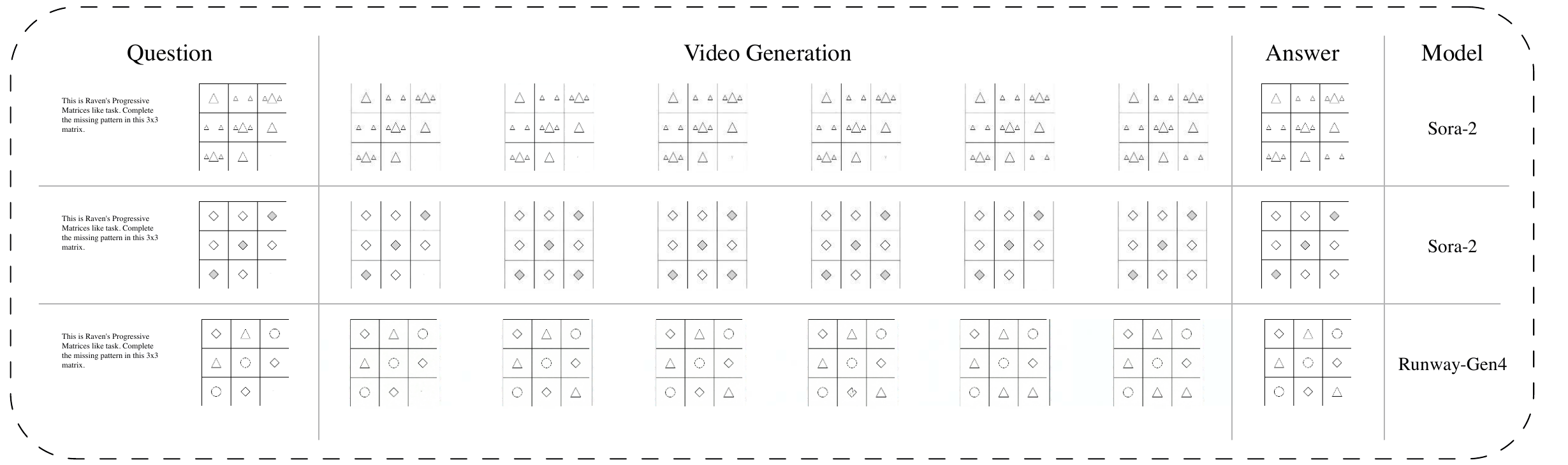}
  \caption{Example runs trying to solve our Raven's Matrices problems. \href{https://grow-ai-like-a-child.com/video-reason/}{Results Page}}
  \label{fig:raven} 
\end{figure*}

The models do exhibit a level of competency in solving our Raven's Matrices problems (see Figure~\ref{fig:raven}). Sora-2 excelled at shape and number progressions but stumbling on rotations and multi-rule combos. Veo 3.0 was strong on color sequences and simple rotations; Veo 3.1 and Runway Gen4-Turbo were middling and sensitive to visual clarity. WaveSpeed WAN 2.2 was weak, and Luma Ray-2 failed entirely (see Figure~\ref{fig:model_comparison}). 

\subsection{Robust Evaluation}

To validate our evaluation methodology, we conducted statistical comparison between GPT-4O assessments and human expert annotations. Statistical analysis reveals exceptionally strong agreement between evaluation methods with Pearson correlation coefficient r = 0.949, indicating near-perfect concordance in scoring patterns. The scatter plot analysis shows consistent linear relationship between automated and human assessments across all score levels (1-5), with point clustering demonstrating frequent agreement on identical score assignments (see Figure~\ref{fig:eval}).

\begin{figure*}[h]
  \centering
  \includegraphics[width=0.4\textwidth]{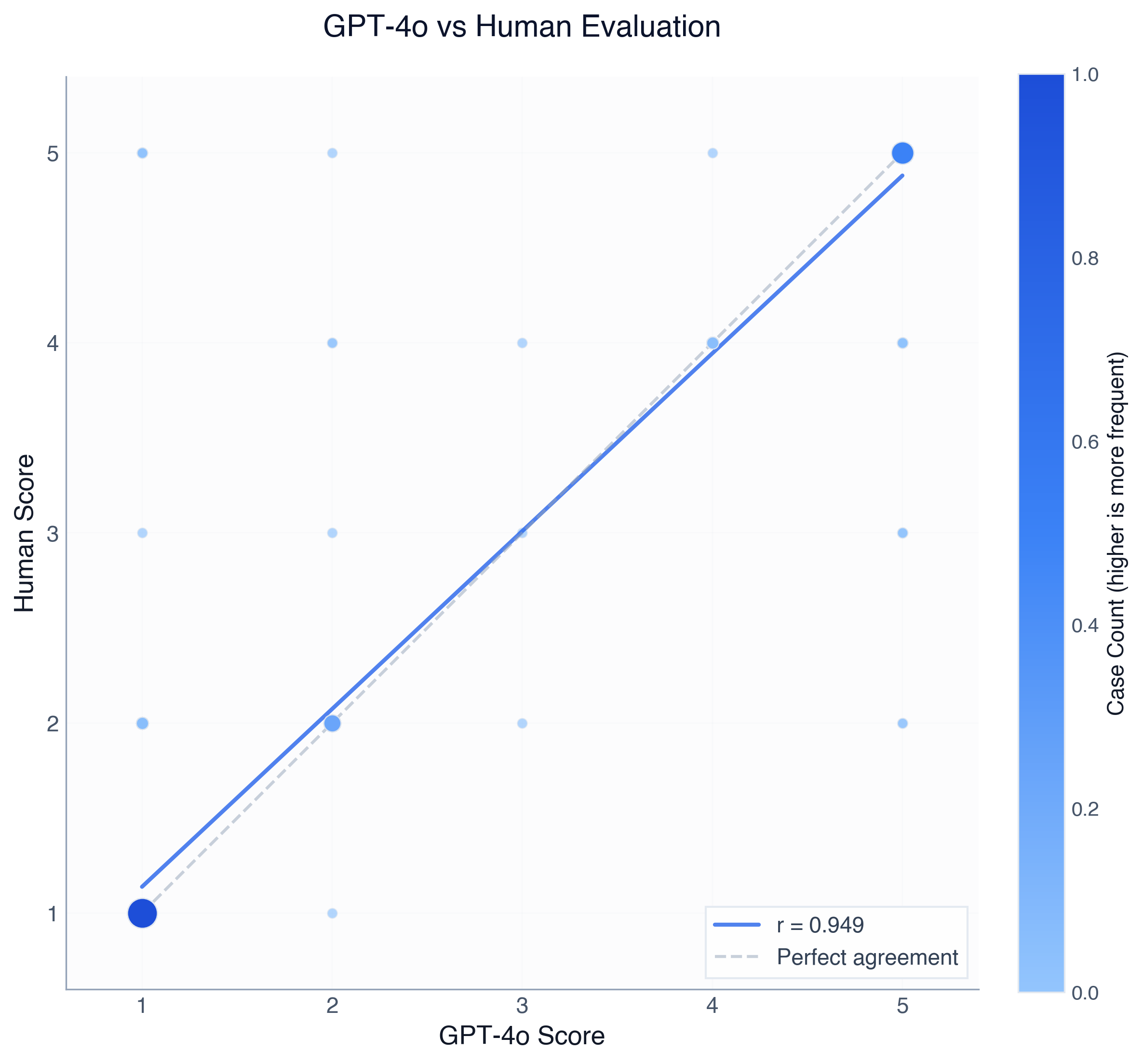}
  \caption{AI Automated versus Human Evaluation}
  \label{fig:eval}
\end{figure*}

\section{Discussion}

Our work adds evidences that scaling laws and the philosophy of emergence start to work in the pixel domain, together with another recent work which reports similar findings in Veo-3 \citep{wiedemer2025video_zero_shot}. Our work extends their efforts by buiding an integrated modular code framework \href{https://github.com/hokindeng/VMEvalKit}{VMEvalKit} which allow users to add models and tasks at scale. The emergence of reasoning in video models is another victory of the scaling hypothesis. It is believed that intelligence comes from compression and therefore in order to achieve intelligence is the path is to scale data, and useful data \citep{sutton2019bitter, sutskever2023observation}. This belief worked in language space, and will probably work in video space. 

However, several interesting yet hard problems are lying in front of us. What is the best way to evaluate reasoning, and particularly language-free reasoning, in video models? Analogical questions have been asked in humans. Could humans without language ability reason? The answer is yes: aphasia patients could reason and solve difficult problems perfectly \citep{fedorenko2024language}. However, counter-arguments exist as most of the aphasia patients lose the abilities of language later in life and a human brain which does not acquire language throughout its entire life is particular rare \citep{penfield1959speech}. Therefore, it seems to be the case that there exists language-free reasoning, but how to find it, as function as such, seems to be particularly difficult both in the human and the machine minds. In current video models, it's obvious that their training data are contaminated with language. 

Not only hard is its ontology, but also its metrology. For human scientists, how to prompt a patient without language ability to understand the task that you want the patient to play. This could be put as the "rule-following" problem for language-free thought \citep{wittgenstein1958philosophical, Kripke1982}. In layman's terms, how do you know both of you are playing the same game when two of you could not communicate via language. On the opposite side, it's could also be put that if the aphasia patient is able to win a competent in a game of chess, it's very unlikely if the patient is not playing chess at all, and not understanding the rule of chess \citep{cappelen2024introspective, cao2025mental, cappelen2025going}. Nevertheless, assessing reasoning and thought without language at all, both in humans and machines, still remain a very hard problem. In our paradigm, we still take the advantage of text prompts to import "rule-following".

Let's say we are able to find existence of language-free reasoning and evaluate it fine. What is its eidos? One interesting paper about hippocampus has shown that human brains use the same substrate for retrieving the past and imagining the future \citep{hassabis2007patients}, while other recent papers argue probably it's also how human conceptual reasoning are done \citep{xiao2025human, veselic2025reasoning}. The central belief behind this line of ideas is that language is not the substrate of the reasoning and thought but very rich experiential content \citep{sim_vs_render_2025}, with some empirical support for and against it \citep{balaban2025capacity, balaban2025physics}. In other words, we could believe that there is such thing that one could reason with only world content, the very rich experiences of senses and actions, which essentially comes from the compression of world itself. 

There are also some engineering opportunities laying in front of us. First, one should be able to do supervised finetuning on the reasoning perceptual flow of the video models. Just like how language models are finetuned on chain-of-thought data, we should be able to display correct reasoning flow, and use that to finetune the reasoning ability in video models. Second, given that we have correctness signal on each task, we should also be able to use reinforcement learning to improve reasoning in video models. Third, understanding the functional decompositions of each ability in the video models offer another opportunity to use video models as experiential intelligence in silico as such, just like how we use language models as models of language \citep{milliere2024language}. Fourth, it also offers tremendous mechanistic interpretability opportunities in understanding what's happening in video models that enable reasoning, just like of language reasoning in language models \citep{wu2025mechanistic, wu2025transformers, sheta2025behavioral}. 

\section{Conclusion}

We present the evidence that video generation models can reason across hard tasks, enabled by our scalable \href{https://github.com/hokindeng/VMEvalKit}{VMEvalKit} framework, which reveals strong performance in top models and paves the way for future advances in evaluation, fine-tuning, and interpretability.

\newpage

\bibliographystyle{plainnat}
\bibliography{references}

\newpage
\appendix
\section{Detailed Task Generation}
\label{sec:detailed_task}

\subsection*{Chess Reasoning - Tactical Pattern Generation}

Our chess module implements a self-contained mate-in-1 generator producing 150+ verified positions without external dependencies. The system employs multiple strategic templates:

\textbf{Back-Rank Mate Generation}: Systematic enumeration of king positions (8 files), pawn structures (20+ configurations), and attacking piece placements. Algorithm generates all combinations of king positions k7, 1k6, 2k5... with corresponding pawn barriers ppp5, 1ppp4, 2ppp3... and attacking pieces R6K, Q6K, 1R5K producing 50+ unique back-rank scenarios.

\textbf{Queen Corner Patterns}: Algorithmic generation of Queen+King endgames with enemy king in corners. System tests 10 queen positions × 4 corner king placements × 5 white king support positions, validating mate moves Qa8\#, Qb8\#, Qc8\# through chess engine verification.

\textbf{Tactical Combination Templates}: Knight mates, rook endgames, and piece coordination patterns with automated validation. Each position undergoes legal move verification and mate confirmation through the python-chess library.

\textbf{Position Validation Pipeline}: All generated positions pass through rigorous validation: (1) FEN string correctness, (2) legal position verification, (3) mate-in-1 confirmation, (4) uniqueness checking via position hashing, (5) visual rendering validation.

\subsection*{Maze Navigation - Graph-Theoretic Generation}

Maze creation employs Kruskal's minimum spanning tree algorithm ensuring unique solution paths:

\textbf{Grid Generation}: 3×3 lattice graphs with systematic edge enumeration and random weight assignment. Kruskal's algorithm produces maze connectivity guaranteeing single solution paths between arbitrary start/end positions.

\textbf{Professional Rendering}: Custom matplotlib-based visualization with precise coordinate systems, consistent styling, and high-resolution output (832×480 pixels, 100 DPI). Green circle markers indicate current position, red flag markers show goals.

\textbf{Difficulty Scaling}: Grid size limitations (3×3 only) ensure consistent complexity while maintaining visual clarity for video model processing. Path length distribution analysis ensures balanced difficulty across generated instances.

\subsection*{Raven Progressive Matrices - Rule-Based Pattern Synthesis}

Our RPM generator implements systematic pattern creation following cognitive psychology principles:

\textbf{Rule Categories}: 
\begin{itemize}
    \item Shape Progression: Geometric transformations (triangle→square→circle)
    \item Number Progression: Quantity changes (1→2→3 objects)
    \item Rotation Patterns: Angular transformations (0°→90°→180°)
    \item Color Sequences: Hue progressions (red→blue→green)
    \item Combination Rules: Multiple simultaneous pattern types
\end{itemize}

\textbf{Matrix Construction}: 3×3 grids with systematic rule application ensuring logical consistency. Bottom-right cell removal creates completion tasks with unique solutions determinable through pattern extrapolation.

\textbf{Visual Rendering}: Custom PIL-based graphics with 150×150 pixel tiles, consistent styling, and clear pattern visibility. Total matrix size 450×450 pixels optimized for model input requirements.

\subsection*{3D Mental Rotation - Voxel Structure Generation}

Spatial reasoning tasks employ sophisticated 3D structure generation with camera manipulation:

\textbf{Voxel Snake Algorithm}: Recursive 3D path generation creating connected cube structures. Algorithm parameters: N=8-9 cubes, segment lengths Lmin=2 to Lmax=5, branching probability p\_branch=0.2, maximum degree constraints preventing overcrowding.

\textbf{Spatial Validation}: Generated structures must span all three coordinate axes (x,y,z) ensuring genuine 3D complexity. Anti-symmetry checks prevent rotationally equivalent structures reducing task difficulty.

\textbf{Camera System}: Professional 3D rendering with controlled viewpoints:
\begin{itemize}
    \item Elevation angles: 20-40° (consistent tilt for 3D visibility)
    \item Azimuth rotations: Exactly 180° horizontal transitions
    \item Perspective projection with equal aspect ratios
    \item Consistent lighting and material properties
\end{itemize}

\textbf{Rendering Pipeline}: Matplotlib 3D with Poly3DCollection faces, consistent coloring (\#7070b0), proper edge definition, and 400×400 pixel output resolution.

\subsection*{Sudoku Logical Reasoning - Constraint Satisfaction Implementation}

3×3 Sudoku generation employs complete enumeration of valid Latin squares:

\textbf{Solution Space}: Pre-computed catalog of all 12 distinct 3×3 Latin square solutions ensuring mathematical completeness. Random selection provides solution diversity while guaranteeing validity.

\textbf{Puzzle Construction}: Systematic number removal (exactly 1 digit) creating unique completion challenges. Difficulty consistency maintained through uniform removal patterns across all generated instances.

\textbf{Visual Presentation}: Clean grid rendering with matplotlib patches, consistent typography (24pt bold), and clear empty cell indication through light gray backgrounds.

\newpage
\section{Failure Modes}
\label{sec:failure}

\subsection{Solving Chess}

\begin{figure*}[h]
  \centering
  \includegraphics[width=1\textwidth]{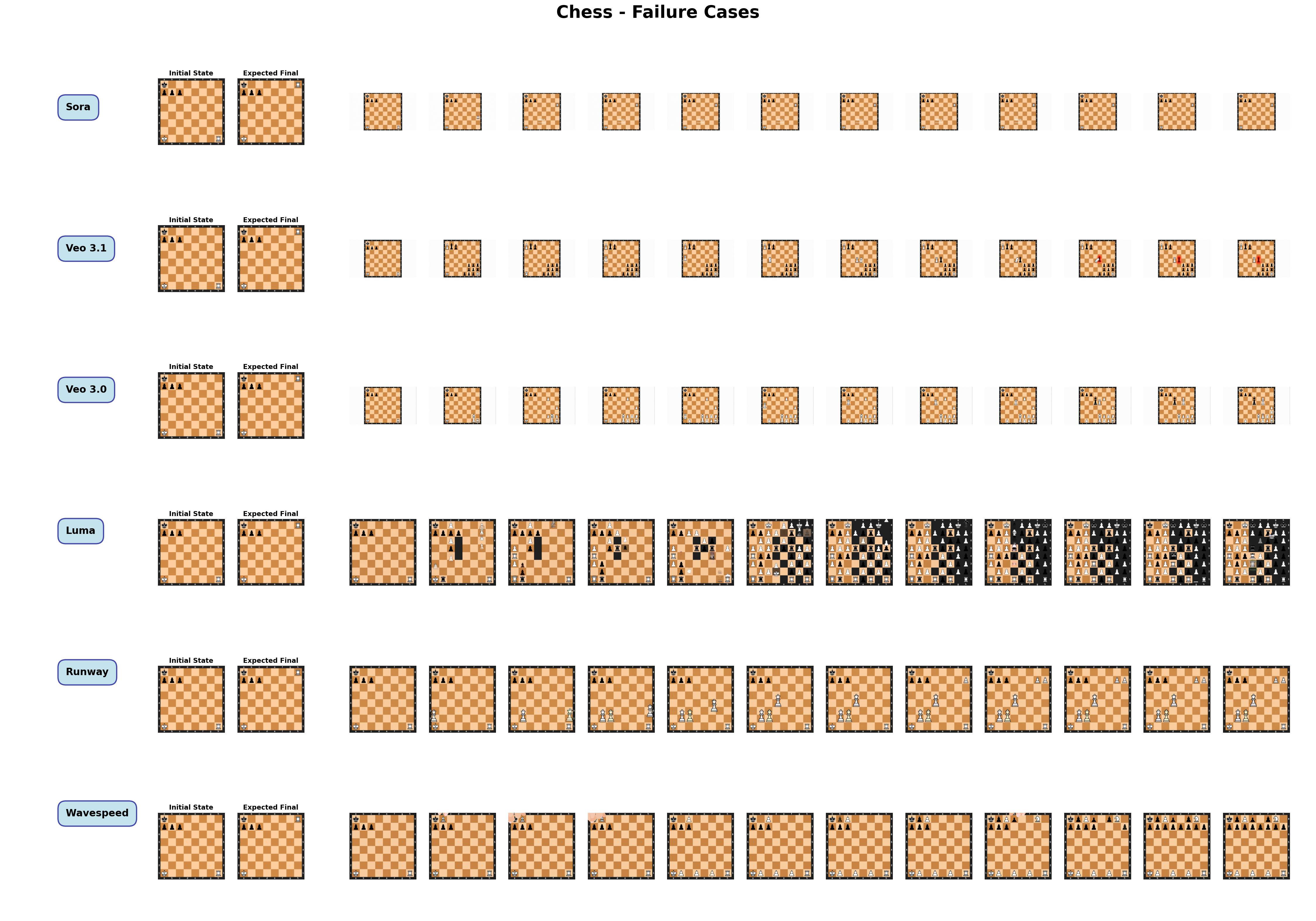}
  \caption{Failure Modes on Chess}
  \label{fig:eval}
\end{figure*}

\newpage

\subsection{Solving Maze}
\begin{figure*}[h]
  \centering
  \includegraphics[width=1\textwidth]{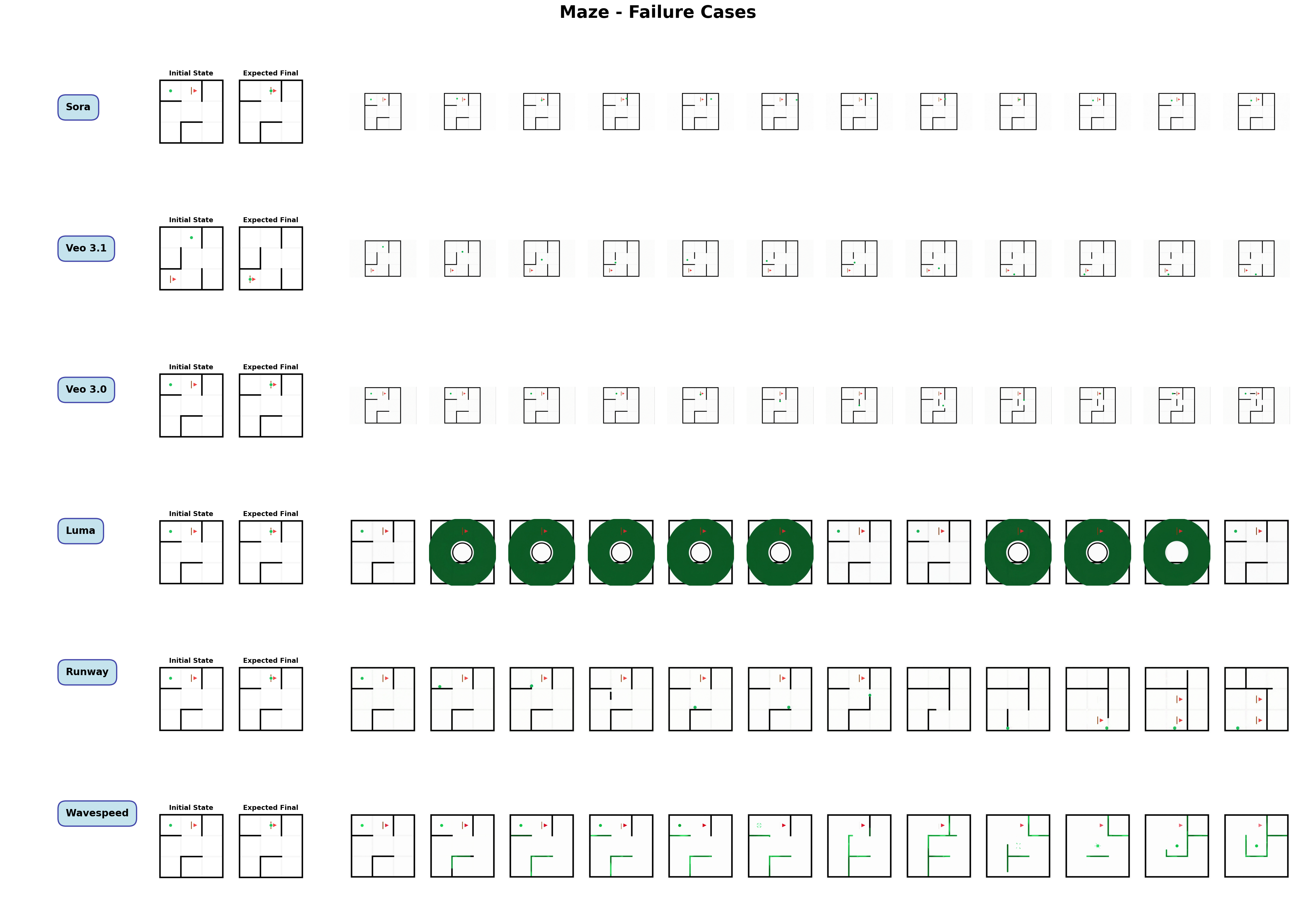}
  \caption{Failure Modes on Maze}
  \label{fig:eval}
\end{figure*}

\newpage

\subsection{Solving Sudoku}
\begin{figure*}[h]
  \centering
  \includegraphics[width=1\textwidth]{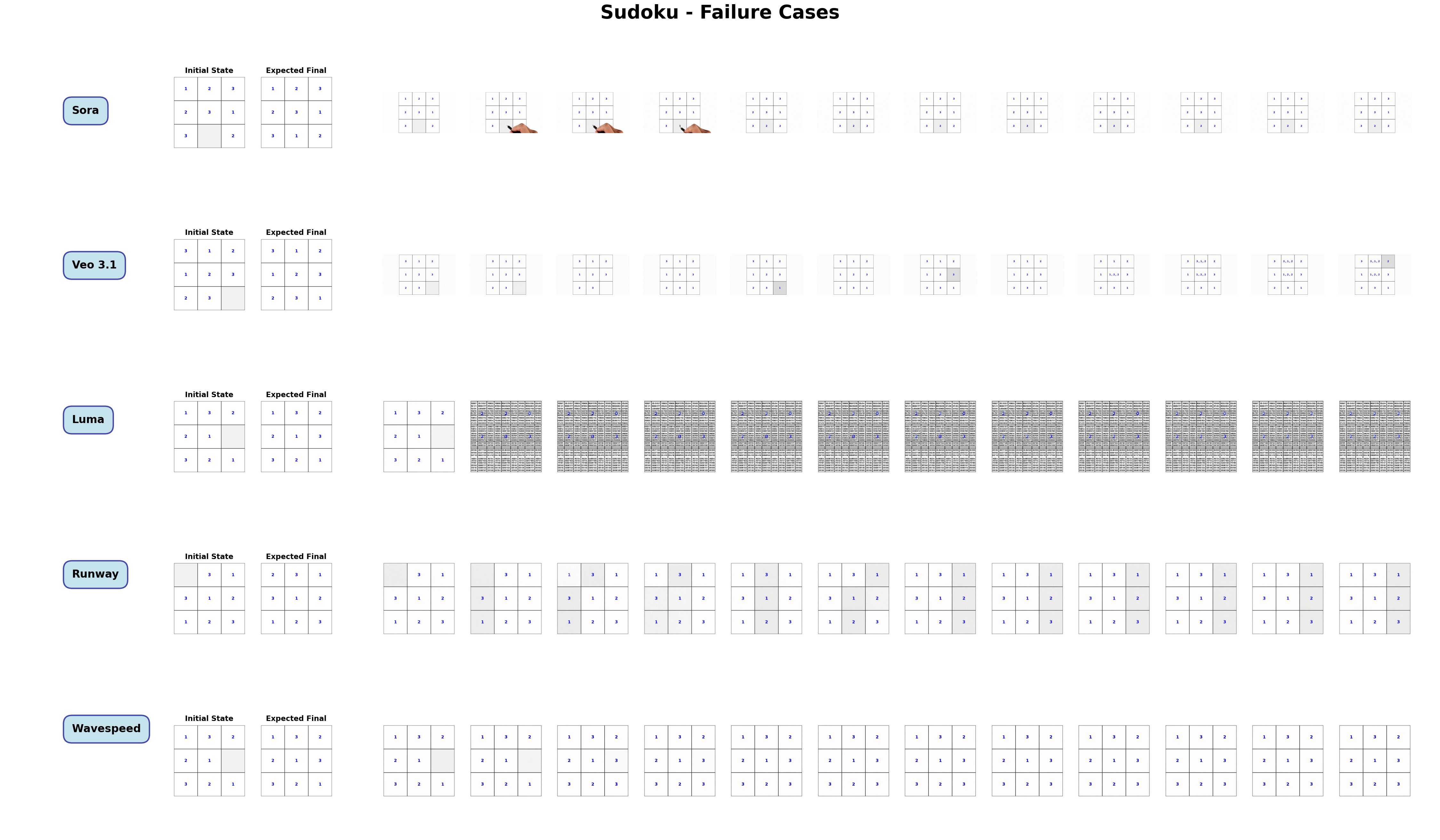}
  \caption{Failure Modes on Sudoku}
  \label{fig:eval}
\end{figure*}

\newpage

\subsection{Solving Mental Rotation}
\begin{figure*}[h]
  \centering
  \includegraphics[width=1\textwidth]{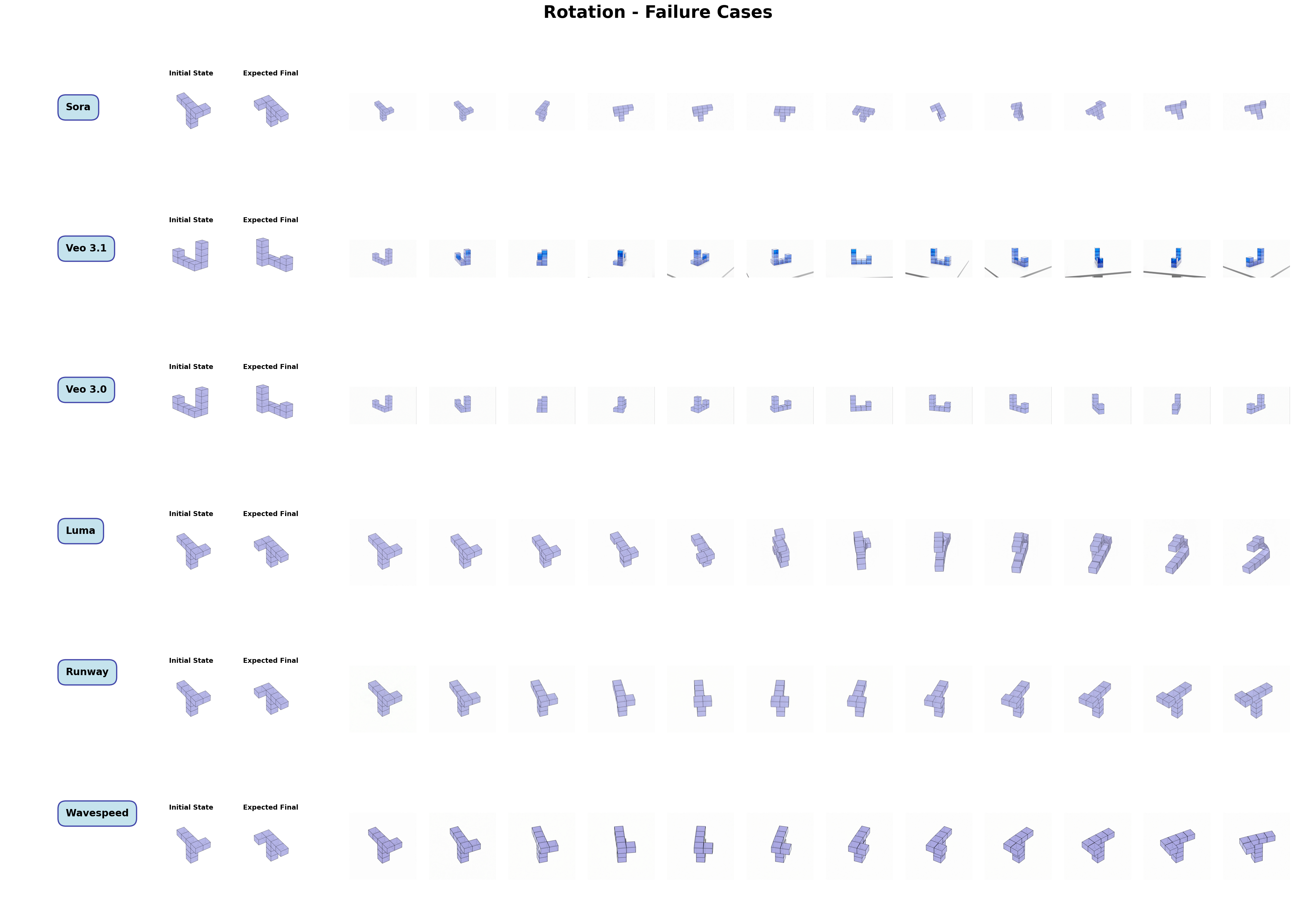}
  \caption{Failure Modes on Mental Rotation}
  \label{fig:eval}
\end{figure*}

\newpage

\subsection{Solving Raven's Matrices}
\begin{figure*}[h]
  \centering
  \includegraphics[width=1\textwidth]{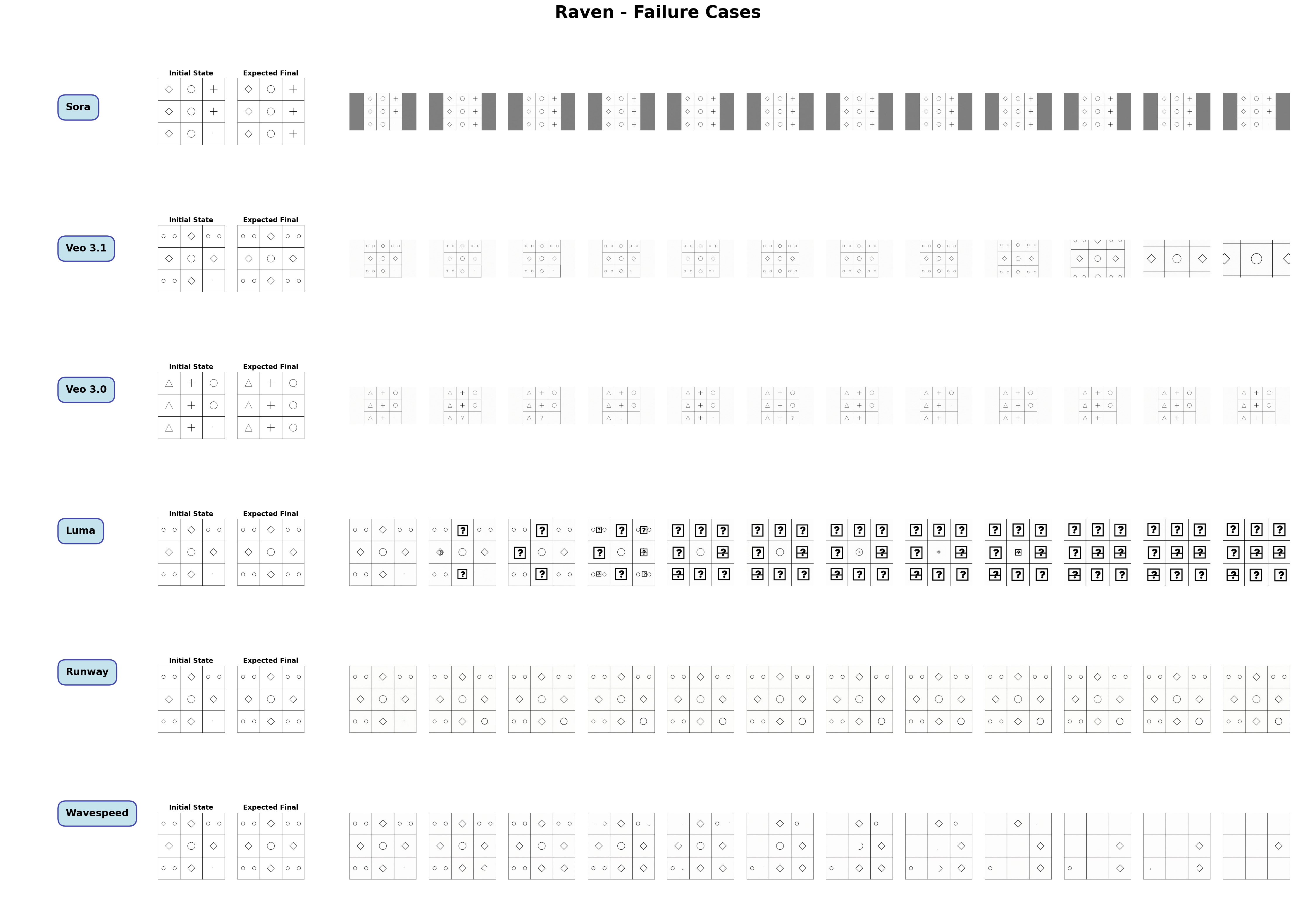}
  \caption{Failure Modes on Raven's Matrices}
  \label{fig:eval}
\end{figure*}





\end{document}